\documentclass[letterpaper]{article} 
\usepackage{aaai2026}  
\usepackage{times}  
\usepackage{helvet}  
\usepackage{courier}  
\usepackage[hyphens]{url}  
\usepackage{graphicx} 
\usepackage{natbib}  
\usepackage{caption} 
\usepackage{algorithm}
\usepackage{algorithmic}
\usepackage{booktabs}
\usepackage{bm}
\usepackage{lmodern}

\usepackage{newfloat}
\usepackage{listings}
\DeclareCaptionStyle{ruled}{labelfont=normalfont,labelsep=colon,strut=off} 
\floatstyle{ruled}
\newfloat{listing}{tb}{lst}{}
\floatname{listing}{Listing}

\title{Bayes-Entropy Collaborative Driven Agents for \\Research Hypotheses Generation and Optimization}

\author {
	Shiyang Duan\textsuperscript{\rm 1}\thanks{Email: astrodsy@163.com},
	Yuan Tian\textsuperscript{\rm 2},
	Qi Bing\textsuperscript{\rm 2},
	Xiaowei Shao\textsuperscript{\rm 1}\thanks{Corresponding author, Email: shaoxwmail@163.com}
}
\affiliations {
	\textsuperscript{\rm 1}School of Aeronautics and Astronautics\
	\textsuperscript{\rm 2}School of Electronic Information and Electrical Engineering\
}

\begin{document}

\maketitle

\begin{abstract}

The exponential growth of scientific knowledge has made the automated generation of scientific hypotheses that combine novelty, feasibility, and research value a core challenge. Existing methods based on large language models fail to systematically model the inherent in hypotheses or incorporate the closed-loop feedback mechanisms crucial for refinement. This paper proposes a multi-agent collaborative framework called \textbf{HypoAgents}, which for the first time integrates Bayesian reasoning with an information entropy-driven search mechanism across three stages-hypotheses generation, evidence validation, and hypotheses Refinement-to construct an iterative closed-loop simulating scientists' cognitive processes. Specifically, the framework first generates an initial set of hypotheses through diversity sampling and establishes prior beliefs based on a composite novelty-relevance-feasibility (N-R-F) score. It then employs etrieval-augmented generation (RAG) to gather external literature evidence, updating the posterior probabilities of hypotheses using Bayes' theorem. Finally, it identifies high-uncertainty hypotheses using information entropy \(H =  - \sum {{p_i}\log {p_i}} \) and actively refines them, guiding the iterative optimization of the hypothesis set toward higher quality and confidence. Experimental results on the ICLR 2025 conference real-world research question dataset (100 research questions) show that after 12 optimization iterations, the average ELO score of generated hypotheses improves by 116.3, surpassing the benchmark of real paper abstracts by 17.8, while the framework's overall uncertainty, as measured by Shannon entropy, decreases significantly by 0.92. This study presents an interpretable probabilistic reasoning framework for automated scientific discovery, substantially improving the quality and reliability of machine-generated research hypotheses.

\end{abstract}


\section{Introduction}
\label{sec:intro}

Scientific progress is fundamentally driven by the generation and refinement of research hypotheses. In the age of large-scale scientific publishing and knowledge explosion, researchers are confronted with an overwhelming volume of information, making it in creasingly difficult to identify meaningful, novel, and testable hypotheses. This challeng has sparked growing interest in the use of artificial intelligence--particularly large language models (LLMs)--to assist or automate parts of the scientific discovery pipeline, including hypotheses generation, evaluation, and iteration~\cite{luo2025LLM4SR}.

In recent years, studies in materials science~\cite{kumbhar2025Hypothesis}, social sciences~\cite{yang2024Large}, biomedicine, and other fields have attempted to leverage LLMs to propose scientific hypotheses. These systems typically employ RAG, multi-turn dialogues, structured prompts, and other methods to generate initial hypotheses, which are then iteratively refined by experts or agent systems~\cite{jinheonbaek2024ResearchAgent,pu2024IdeaSynth}. However, most existing approaches are limited to one-time generation or shallow optimization, lacking a closed-loop simulation. They also lack systematic mechanisms for hadndling uncertainty~\cite{li2024Learning,xianghu2024Nova}.

To address the aforementioned challenges, this paper proposes a novel multi-agent framework, termed HypoAgents. Centered around a "Propose-Verify-Refine" closed-loop, the method incorporates Bayesian inference and an information entropy-driven mechanism to simulate the exploratory behavior of real-world researchers in knowledge-incomplete enviroments. The core research question we focus on is: Under the permise of limited evidence, how can intelligent agents systematically propose, evaluate, and optimize a collection of research hypotheses that meet high-quality requirements?

Specifically, the proposed HypoAgents framework consists of three core stages: (1) \textbf{Hypothesis Proposal}: generating an initial hypothesis set through diversity sampling and semantic clustering; (2) \textbf{Evidence Validation}: constructing hypothesis-evidenc pairs based on RAG retrieval and using LLMs as probabilistic evaluators for likelihood scoring; (3) \textbf{Hypothesis Refinement}: idenfifying high-uncertainty hypotheses according to Bayesian update rules and information entropy reduction criteria, followed by targeted modification, thereby gradually converging toward a high-quality hypothesis set.

The innovations of this paper are as follows:

\begin{itemize}
	\item Proposes a novel research hypothesis optimization method that combines Bayesian belief updating with an entropy-driven search mechanism, effectively balancing exploration and convergence. This method ensures the continuous refinement of hypotheses through an iterative process that mimics scientific reasoning.
	\item Introduces Bayesian inference and uncertainty analysis for the first time as guiding principles in the iterative optimization of research hypotheses. This enhances the reliability of the hypothesis generation process, making it more robust in handling the uncertainties inherent in scientific discovery.
	\item The efficacy of the proposed method is demonstrated through a large-scale evaluation on the ICLR 2025 conference research question dataset. Results reveal a significant improvement, with a 116.3-point increase in the average ELO score for generated hypotheses after 12 optimization iterations, surpassing real paper abstracts by 17.8 points. Moreover, the uncertainty associated with the hypotheses decreases by 0.92, indicating a stronger confidence in the generated hypotheses.
\end{itemize}

The structure of this paper is as follows: Section 2 reviews related work on LLM-driven hypothesis generation. Section 3 details the proposed HypoAgents framework, including the three core stages of hypothesis proposal, evidence validation, and refinement. Section 4 describes the experimental setup and evaluation metrics. Section 5 presents and analyzes the results of the experiments, highlighting the effectiveness of our approach. Finally, Section 6 concludes the paper and discusses future research directions.

\section{Related Works}
\label{sec:related}

\subsection{LLM-Driven Hypothesis Generation}

In recent years, LLMs have shown significant potential in generating scientific hypotheses and research ideas, primarily achieved through various prompt engineering and interactive paradigms for automated ideation. We categorize the existing literature into two main strategies: \textbf{single-model prompting} and \textbf{multi-round iterative generation}.

\subsubsection{Single-Model Prompting}

Early research primarily enhanced the reasoning capabilities of LLMs through Chain-of-Thought (CoT) prompting or RAG. For example, MOOSE-Chem employs a simple RAG pipeline, enabling an LLM to rediscover previously published chemistry research hypotheses from top-tier journals without domain-specific fine-tuning~\cite{yang2024MOOSEChem}. Similarly, the Chain-of-Ideas (CoI) framework links retrieved concepts into a structured "thought chain," guiding the LLM to generate more novel and scientifically profound research ideas~\cite{longli2024Chain}.

Although these methods offer advantages such as high efficiency and low computational cost, their single-round generation nature often results in a lack of self-reflection and correction mechanisms. The generated hypotheses may contain factual errors or lack depth. Furthermore, these approaches typically lack a systematic evaluation of novelty, feasibility, or scientific value.

\subsubsection{Multi-Turn Interative Generation}

To overcome these limitations, recent studies have proposed multi-stage generation frameworks that introduce structured feedback loops to enhance the quality of the output. For example, \textit{ResearchAgent} utilizes a "writer-reviewer" dual-agent loop, where the "writer" is responsible for idea generation and the "reviewer" provides critical feedback, significantly improving the novelty and relevance of the results~\cite{jinheonbaek2024ResearchAgent}. \textit{IdeaSynth} decomposes research hypotheses into reusable "facets" (problem, method, evaluation) and conducts structured exploration through evolution and recombination~\cite{pu2024IdeaSynth}.

Furthermore, frameworks like \textit{Nova} and \textit{Learn2Gen} have introduced more complex planning and evaluation mechanisms. \textit{Nova} uses a "Plan-Retriever-Search" cycle to enhance knowledge integration and generation diversity~\cite{xianghu2024Nova}, while \textit{Learn2Gen} combines supervised fine-tuning with reinforcement learning based on multi-objective rewards to effectively balance multiple dimensions such as novelty, feasibility, and clarity~\cite{li2024Learning}.

These approaches mark a shift from static prompting to dynamic, feedback-driven generation models, advancing the evolution of LLMs toward scientific intelligent agents with reasoning capabilities.

\subsection{Multi-Agent Collaboration and Automated Scientific Workflows}

Beyond individual generation strategies, another significant direction is the construction of multi-agent collaborative frameworks to support the automation of the scientific process, including stages like ideation, experimental design, and paper writing.

The \textit{AI-Scientist} framework is representative in this domain, achieving full-process automation by combining multiple agents with specific decision-making capabilities (such as ideation, code execution, experimental evaluation, and paper writing)~\cite{chrislu2024AI}. Its second-generation framework,\textit{ AI-Scientist-v2}, introduces tree search-based experimental planning and a vision-language model closed-loop, achieving fully automated and peer-reviewed research outputs~\cite{yamada2025AI}. Similarly, the \textit{SciAgents} framework coordinates heterogeneous agents (e.g., Ontologist, Explorers, Critics) to collaborate on a dynamic knowledge graph, generating and refining interdisciplinary hypotheses through semantic path sampling and inter-agent evaluation~\cite{alirezaghafarollahi2024SciAgents}.

The effectiveness of these frameworks relies on two key mechanisms: (1) \textbf{division of labor}, where agents have clear responsibilities in phases like generation, validation, or planning (e.g., AI-Scientist and Nova); and (2) structured critique and revision, typically implemented through the collaboration of multiple peer-evaluating agents (e.g., ResearchAgent, Learn2Gen).

Although the existing methods have demonstrated the potential of LLMs in research ideation and workflow collaboration, most still have shortcomings in the following areas: integrating belief updating, evaluating evidence-based support, and navigating uncertain hypothesis spaces.

Building on prompt design, multi-turn feedback, and multi-agent collaboration, our work integrates a probabilistic reasoning framework to construct a unified and interpretable LLM-driven scientific hypothesis generation scheme.

\section{Methodology}

This section presents a detailed description of our proposed framework for automatic generation and iterative refinement of research hypotheses, termed \textbf{HypoAgents}. The framework is designed to simulate the core "propose-validate-refine" loop in scientific research. It systematically generates, evaluates, and refines scientific hypotheses through multi-agent collaboration.

\subsection{Problem Definition}

The central task of this research is to automatically generate a high-quality set of research hypotheses $H=\left\{h_1,h_2,\dots,h_n\right\}$ for a given open-ended research question $Q$. A high-quality hypothesis must strike an optimal balance across three core dimensions:

\begin{enumerate}
	\item \textbf{Novelty} (\(N\)): Does the hypothesis propose new insights, mechanisms, or associations that extend beyond the prevailing knowledge in the literature?
	\item \textbf{Relevance} (\(R\)): Is the hypothesis closely aligned with the core research question $Q$, and can it help address the question either directly or indirectly?
	\item \textbf{Feasibility} (\(F\)): Is the hypothesis testable in the real world through experimentation, data analysis, or other scientific methods?
\end{enumerate}

This task can be formalized as a multi-objective optimization problem: finding an optimal hypothesis set $H^*$ that maximizes a composite evaluation function $\mathcal{J} \left(H;N,R,F\right)$. Our framework approaches this goal through an iterative Bayesian inference process.

\subsection{Framework Overview}

The \textbf{HypoAgents} framework consists of three core modules: \textbf{Hypothesis Proposal}, \textbf{Evidence Validation}, and \textbf{Hypothesis Refinement}. These modules form a closed-loop optimization framework. Starting with a research question $Q$, the framework first generates a batch of diverse initial hypotheses. Next, it retrieves evidence from external knowledge sources to validate them, and updates its belief in each hypothesis using Bayes’ theorem. It then identifies hypotheses with high uncertainty or weak support and applies targeted revisions, entering the next “validation-revision” cycle. This process continues until convergence.

\subsection{Hypothesis Proposal}

The goal of this stage is to generate a diverse and initially plauisible set of hypotheses $H_0 = \left\{h_1,h_2,\dots,h_n\right\}$ for the research question \(Q\) and assign prior probabilities to them.

\subsubsection{Diverse Hypothesis Generation}

To ensure breadth and diversity in the initial hypothesis set, we adopt a two-stage strategy:

\begin{enumerate}
	\item \textbf{Multi-round Sampling with LLMs}: Leveraging the generative capacity of large language models (LLMs), we perform multi-round sampling by varying the temperature parameter and designing diverse prompt templates. This allows the exploration of different angles of the research question to generate a large pool of candidate hypotheses.
	\item \textbf{Semantic Clustering and Selection}: To filter representative and non-redundant hypotheses from the large candidate pool, we first map each hypothesis into a high-dimensional vector space using a pretrained embedding model. Then, we apply the K-Means clustering algorithm to group semantically similar hypotheses. To ensure representativeness and avoid redundancy, we then select the hypothesis closest to the centroid of each cluster to form the initial hypothesis set $H_0$.
\end{enumerate}

\subsubsection{Initial Belief Construction}

To initiate the Bayesian iterative process, each hypothesis $h_i \in H_0$ is assigned an initial belief as its prior probability. We define an initial belief score $B_0 \left(h_i\right)$, which is a normalized weighted sum incorporating novelty, relevance, and feasibility:

$$B_0 \left(h_i\right) = \frac{{\alpha  \cdot N\left( {{h_i}} \right) + \beta  \cdot R\left( {{h_i}} \right){\rm{ + }}\gamma  \cdot F\left( {{h_i}} \right)}}{{\sum\limits_{j = 1}^n {\left( {\alpha  \cdot N\left( {{h_j}} \right) + \beta  \cdot R\left( {{h_j}} \right){\rm{ + }}\gamma  \cdot F\left( {{h_j}} \right)} \right)} }}$$

Here, $N\left(h_i\right), R\left(h_i\right), F\left(h_i\right) \in \left[0,1\right]$ are scores assigned to each hypothesis by an LLM-based evaluator. The hyperparameters $\alpha, \beta, \gamma$ represent the importance weights for novelty, relevance, and feasibility respectively, and satisfy $\alpha + \beta + \gamma = 1$. The weighted scores are normalized to form a probability distribution $B_0 = \left\{B_0\left(h_1\right), \dots, B_0\left(h_n\right)\right\}$, representing the prior belief over the hypothesis set $H_0$.

\subsection{Evidence Validation}
\label{sec:Evidence Validation}

This stage centers on objectively evaluating the validity of each hypothesis based on external knowledge and updating the belief accordingly using Bayes’ theorem.

\subsubsection{Literature-Based Evidence Retrieval}

We construct a knowledge base composed of domain-specific academic literature. For each hypothesis $h_i$ under evaluation, the framework forms a structured query by concatenating it with the research question $Q$, and retrieves the top-$k$ most relevant text segments from the vectorized knowledge base to form the evidence set $D_i = \left\{d_1, d_2, \dots, d_m\right\}$.

\subsubsection{Likelihood Estimation}

The likelihood function $L\left(D_i | h_i\right)$ quantifies the probability of observing the evidence set $D_i$ assuming that the hypothesis $h_i$ is true. We design a dual-evidence evaluation mechanism to assess each piece of evidence $d_j \in D_i$:

\begin{itemize}
	\item \textbf{Base Likelihood Score} $L_{base} \left(d_j | h_i\right)$: We use an LLM as a probabilistic evaluator to estimate $P\left({{d_j}|{h_i}}\right)$. Specifically, we prompt the model with: "Assume the following scientific hypothesis is true: {\(h_i\)}. How likely is it that the following piece of literature evidence would be observed: {\(d_j\)}? Please give a continuous score between 0 and 1, where 0 means highly unlikely and 1 means fully consistent." The model outputs a direct probability estimate.
	\item \textbf{Methodological Alignment Score} $M \left(d_j, h_i\right)$: We use an LLM as a binary classifier to determine whether \(d_j\) includes methodological elements (e.g., experimental design, analytical approaches) that support or test \(h_i\). If so, $M \left(d_j, h_i\right) = 1$; otherwise, it is 0.
\end{itemize}

The final contribution of each piece evidence is computed as:

\[L\left( {{d_j}|{h_i}} \right) = {L_{base}}\left( {{d_j}|{h_i}} \right) \cdot M\left( {{d_j},{h_i}} \right)\]

To mitigate the impact of low-quality individual evidence (e.g., from LLM errors), we use an average aggregation rather than multiplicative accumulation for the total likelihood:

\[L\left( {{D_i}|{h_i}} \right) = \frac{1}{m}\sum\limits_{j = 1}^m {L\left( {{d_j}|{h_i}} \right)} \]

\subsubsection{Bayesian Posterior Update}

Given the computed likelihoods, we update the posterior belief $B_k \left(h_i\right)$ for each hypothesis using Bayes' rule:

\[{B_k}\left( {{h_i}} \right) = \frac{{L\left( {{D_i}|{h_i}} \right) \cdot {B_{k - 1}}\left( {{h_i}} \right)}}{{\sum\nolimits_{j = 1}^n {L\left( {{D_j}|{h_j}} \right) \cdot {B_{k - 1}}\left( {{h_j}} \right)} }}\]

Here, $B_{k-1} \left(h_i\right)$ is the prior belief from the previous iteration (or the initial prior $B_0 \left(h_i\right)$ when \(k=1\)). The denominator serves as a normalization constant ensuring that all posterior beliefs sum to one.

\subsubsection{Uncertainty Metrics}

To monitor convergence, we introduce Shannon entropy to measure the overall uncertainty of the belief distribution. Higher entropy indicates greater uncertainty across hypotheses.

\[{H_k} =  - \sum\limits_{i = 1}^n {{B_k}\left( {{h_i}} \right){{\log }_2}{B_k}\left( {{h_i}} \right)} \]

\subsection{Hypothesis Refinement}
\label{sec:Hypothesis Refinement}

The goal of this stage is to identify and refine hypotheses with the hightest uncertainty to actively explore the hypothesis space.

\subsubsection{Selection for Refinement}

We use \textbf{binary entropy} to quantify the individual uncertainty of each hypothesis \(h_i\):

\[{S_k} =  - {B_k}{\log _2}{B_k} - \left( {1 - {B_k}} \right){\log _2}\left( {1 - {B_k}} \right)\]

This score, representing the individual uncertainty of a hypothesis, peaks when its belief \({B_k}\left( {{h_i}} \right)\) is close to 0.5. This state of maximal uncertainty signals that the accumulated evidence is equally balanced for and against the hypothesis, making it a prime candidate for refinement.

\subsubsection{Refinement Strategies}

For each selected hypothesis, we apply one of several predefined heuristic strategies based on its current state and evidence:

\begin{itemize}
	\item \textbf{Deepening}: If a hypothesis has moderate support but is vaguely formulated, the framework uses an LLM to increase specificity by adding mechanisms, boundary conditions, or scope.
	\item \textbf{Counterfactual}: If a hypothesis is strongly contradicted by evidence, a counter or alternative hypothesis is generated, e.g., by negating its causal claim or proposing a different explanatory framework.
	\item \textbf{Hybridization}: If multiple uncertain hypotheses converge on a similar theme or direction, their core elements (e.g., problem, method, perspective) are recombined to form a more comprehensive and precise new hypothesis.
\end{itemize}

This process is formalized as a refinement function ${\cal R}\left( {{h_i}, D_i, {\rm{strategy}}} \right) \to {h_i}'$, which produces a revised hypothesis \(h_i'\). The refined hypothesis replaces the original and enters the next iteration.

\subsection{Iteration and Termination}

The framework takes the posterior beliefs \(B_k\) from the previous round as the new prior \(B_{k-1}\) and repeats the "Evidence Validation" and "Hypothesis Refinement" stages in an iterative loop. The loop terminates when any of the following conditions are met:

\begin{enumerate}
	\item \textbf{Entropy Convergence}: The change in entropy between two consecutive rounds is less than a preset threshold $\varepsilon _H$, i.e., $\left| {{H_k} - {H_{k - 1}}} \right| < {\varepsilon _H}$.
	\item \textbf{Maximum Iterations Reached}: The number of iterations $k$ reaches a predefined limit $T_{max}$.
\end{enumerate}

Upon termination, the framework outputs the final belief distribution \(B_T\) and the corresponding set of hypotehses \(H_T\) as the final result.

\section{Experiments}
\label{sec:exp}

This section aims to evaluate the effectiveness of the proposed \textbf{HypoAgents} framework. We introduce the datasets, task settings, and core evaluation metrics used in the experiments. Then, we demonstrate and analyze the impact of different hyperparameter configurations on model performance to validate the robustness of the framework and identify optimal practices.

\subsection{Dataset and Task Setup}

\paragraph{Data Source} The experimental data comes from the publicly available paper collection of the ICLR 2025 conference. The conference received a total of 11,672 submissions, of which 3,708 were accepted.
\paragraph{Research Question Construction} From the Top-100 high-scoring papers of ICLR 2025, we automatically extracted and manually selected 100 representative research questions as the starting point for the experimental tasks using LLM. These questions cover a range of cutting-edge topics in machine learning and exhibit high complexity and openness.
\paragraph{Knowledge Base Construction} To simulate a realistic research environment, we constructed a dedicated knowledge base for each research question. Specifically, we parsed the reference lists from the aforementioned Top-100 papers and matched them with the Semantic Scholar database, ultimately obtaining the full text of 928 open-source reference papers. These papers were processed and built into a vectorized knowledge base for retrieving relevant evidence during the hypothesis validation phase.

\subsection{Evaluation Metrics}

To comprehensively assess the performance of HypoAgents, we use the following quantitative metrics:

\paragraph{ELO Score} The ELO rating system, originally developed for ranking players in competitive games, is adapted here to quantitatively evaluate the quality of hypotheses generated across different iterations relative to real paper abstracts. Specifically, we use LLM as a reviewer to perform pairwise comparisons between the hypotheses \(h\) generated in different iterations for the same research question and the actual published paper abstract \(\hat h\). The LLM judges the superior hypothesis in each pairwise comparison. Based on the pairwise comparison results, each hypothesis is assigned a dynamically updated ELO score. In each iteration, the average ELO score of all generated hypotheses is compared to the ELO score of the real paper abstract, and the difference serves as a key performance metric. The change in the ELO score (ELO difference) is used to measure the optimization effect of the hypotheses during each iterative round.
\paragraph{Entropy} As defined in the methodology section, the information entropy \(H\) is used to measure the uncertainty of the entire hypothesis belief distribution. Ideally, during the optimization process, the system should confirm high-quality hypotheses and eliminate low-quality ones through evidence, leading to a steady decrease in the overall entropy value. Therefore, the change in entropy (entropy difference, \(\Delta H\)) is used as a metric to assess the system's convergence and the reduction in uncertainty.

\section{Result and Analysis}
\label{sec:results}

This section analyzes the performance of the proposed HypoAgents framework under different hyperparameters to validate its effectiveness and robustness. The experiments focus on three key hyperparameters: \textbf{Number of Iterations} \(T\), \textbf{Number of Hypotheses} \(n\), and \textbf{Refinement Threshold} \(\tau_s\).

\paragraph{The Number of Iterations \(T\)}

\begin{figure}[ht]
	\centering
	\includegraphics{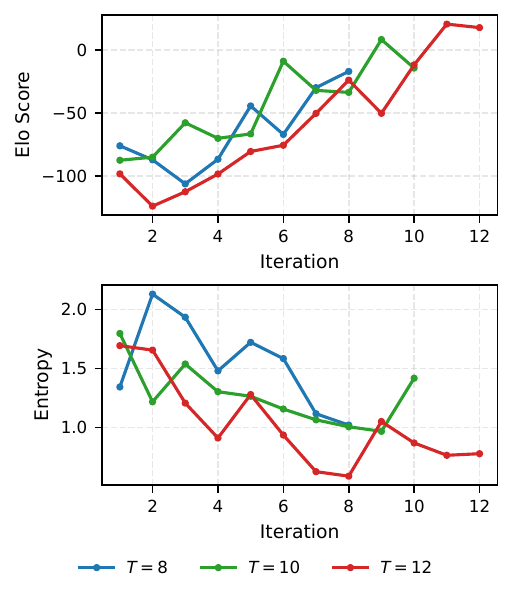}
	\caption{Impact of Different Iterations on Performance}
	\label{fig:sensitivity_T}
\end{figure}

We first analyze the impact of the number of iterations \(T\) on the optimization effect. In the experiments, the refinement threshold is fixed at \(\tau_s=0.3\), and the number of hypotheses is set to \(n=5\), with \(T\) being varied between 8, 10, and 12 for comparison. The experimental results are shown in Table~\ref{tab:sensitivity_T}, and the detailed iterative evolution process is illustrated in Figure~\ref*{fig:sensitivity_T}.

As seen in Table~\ref{tab:sensitivity_T}, the optimization effect of hypotheses significantly improves as the number of iterations \(T\) increases. Specifically, when the number of iterations is increased from 8 to 10, the ELO difference improves from 59.17 to 73.33, indicating that increasing the number of iterations helps enhance the quality of the hypotheses. More notably, when the iterations are further increased to 12, the ELO difference rises to 116.27, with the final round ELO reaching 17.77, representing a positive leap, indicating a significant improvement in the overall quality of the hypotheses.

Furthermore, as the number of iterations increases, the downward trend in entropy becomes more pronounced. When the number of iterations reaches 12, the entropy difference is -0.92, showing a significant reduction compared to 8 and 10 iterations. This suggests that as the number of iterations increases, the system's certainty about the hypothesis distribution continuously strengthens, making the optimization process more robust and effective.

\begin{table}[ht]
	\centering
	\caption{Results Comparison for Different Iterations}
	\label{tab:sensitivity_T}
	\resizebox{\linewidth}{!}{
	\begin{tabular}{lcccc}
		\toprule
		\textbf{\(T\)} & \textbf{First Round ELO} & \textbf{Final Round ELO} & \textbf{ELO \(\Delta\) ↑} & \textbf{\(\Delta H\)}\\
		\midrule
8	&	-76.21	&	-17.04	&	59.17	&	-0.32 \\
10	&	-87.71	&	-14.38	&	73.33	&	-0.38 \\
12	&	-98.50	&	17.77	&	\textbf{116.27}	&	-0.92 \\
		\bottomrule
	\end{tabular}}
\end{table}

\subsubsection{The Number of Hypotheses \(n\)}

\begin{figure}[ht]
	\centering
	\includegraphics{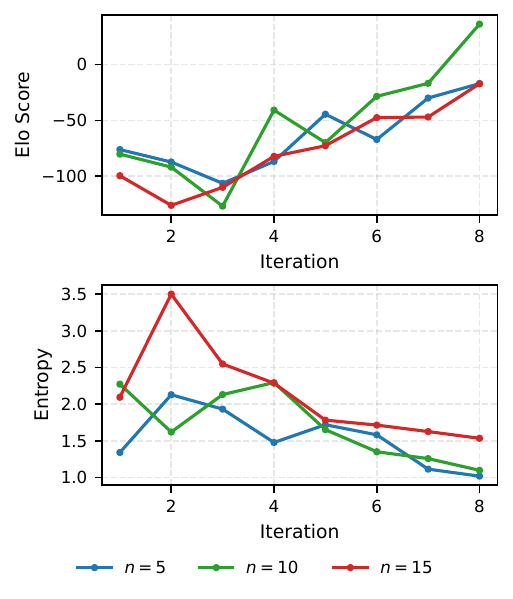}
	\caption{Impact of Different Numbers of Hypotheses on Performance}
	\label{fig:sensitivity_N}
\end{figure}

Next, we analyze the impact of the number of hypotheses \(n\). In this experiment, the number of iterations is fixed at \(T=8\) and the refinement threshold at \(\tau_s=0.3\), with the number of hypotheses set to 5, 10, and 15, respectively. The results are shown in Table~\ref{tab:sensitivity_N}, with the specific optimization results depicted in Figure~\ref{fig:sensitivity_N}.

From Table~\ref{tab:sensitivity_N}, we can see that an appropriate increase in the number of hypotheses \(n\) significantly improves the optimization effect. This suggests that a broader initial search space (a larger \(n\)) provides richer material for the iterative refinement process, increasing the probability of discovering and converging on a high-quality hypothesis. When the number of candidates per round is increased from 5 to 10, the ELO improvement increases from 59.17 to 116.60, and the final round ELO rises from -17.04 to 36.34, reflecting a higher quality improvement. However, when the number of hypotheses is further increased to 15, the ELO improvement decreases, and the final round ELO drops to -17.16. This indicates that there is an optimal range for the number of hypotheses, and too many candidates may introduce redundant or low-quality hypotheses, weakening the optimization effect. An excessive number of candidates (\(n=15\)) may introduce noise and redundancy. This could dilute the focus of the optimization process, spreading the evidence-gathering and thus hindering convergence towards the best hypotheses.

\begin{table}[ht]
	\centering
	\caption{Results Comparison for Different Number of Hypotheses}
	\label{tab:sensitivity_N}
	\resizebox{\linewidth}{!}{
	\begin{tabular}{lcccc}
		\toprule
		\textbf{\(n\)} & \textbf{First Round ELO} & \textbf{Final Round ELO} & \textbf{ELO \(\Delta\) ↑} & \textbf{\(\Delta H\)}\\
		\midrule
5	&	-76.21	&	-17.04	&	59.17	&	-0.32 \\
10	&	-80.27	&	36.34	&	\textbf{116.60}	&	-1.17 \\
15	&	-99.65	&	-17.16	&	82.49	&	-0.56 \\
		\bottomrule
	\end{tabular}}
\end{table}

\subsubsection{The Refinement Threshold \(\tau_s\)}

\begin{figure}[ht]
	\centering
	\includegraphics{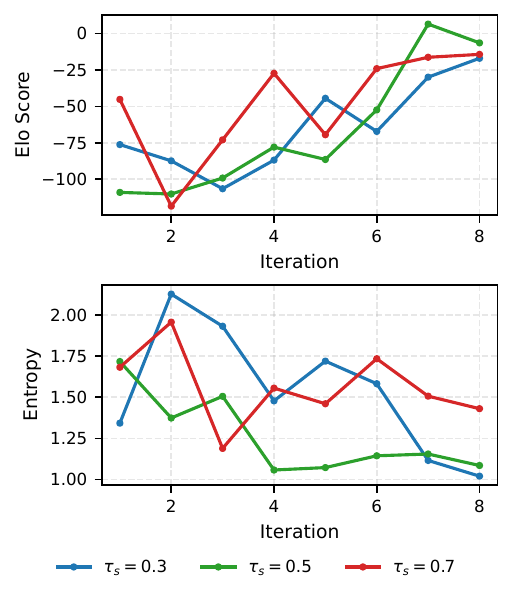}
	\caption{Impact of Different Refinement Thresholds on Performance}
	\label{fig:sensitivity_tau}
\end{figure}

Finally, we investigate the role of the refinement threshold \(\tau_s\). In this experiment, the total number of iterations is fixed at \(T=8\), and the number of hypotheses is set to \(n=5\), with the refinement threshold \(\tau_s\) varied at 0.3, 0.5, and 0.7. The experimental results are shown in Table~\ref{tab:sensitivity_tau}, and the specific optimization results are illustrated in Figure~\ref{fig:sensitivity_tau}.

From Table~\ref{tab:sensitivity_tau} and Figure~\ref{fig:sensitivity_tau}, we observe that when the \(\tau_s=0.5\), the ELO difference increasing by 102.55 and a significant reduction in entropy by 0.63. In contrast, when the threshold is set too low (e.g., \(\tau_s=0.3\)), the selection criteria for refinement become too permissive. This leads to the frequent modification of hypotheses that already have relatively low uncertainty, limiting the exploration of more genuinely ambiguous but potentially valuable ideas. On the other hand, when the threshold is set too high (e.g., \(\tau_s=0.7\)), the selection becomes overly stringent, leading to the premature elimination of potential high-quality hypotheses, which limits the overall exploration ability of the algorithm and results in the smallest performance improvement, just 30.87.

\begin{table}[ht]
	\centering
	\caption{Results Comparison for Different Refinement Thresholds}
	\label{tab:sensitivity_tau}
	\resizebox{\linewidth}{!}{
	\begin{tabular}{lcccc}
		\toprule
		\textbf{\(\tau_s\)} & \textbf{First Round ELO} & \textbf{Final Round ELO} & \textbf{ELO \(\Delta\) ↑} & \textbf{\(\Delta H\)}\\
		\midrule
0.3	&	-76.21	&	-17.04	&	59.17	&	-0.32 \\
0.5	&	-109.00	&	-6.46	&	\textbf{102.55}	&	-0.63 \\
0.7	&	-45.23	&	-14.36	&	30.87	&	-0.25 \\
		\bottomrule
	\end{tabular}}
\end{table}

\section{Limitations and Future work}

Although HypoAgents shows encouraging results, several limitations remain.

\begin{enumerate}
	\item \textbf{Dynamic Evidence Integration}: The current knowledge base is static, relying on a pre-compiled literature snapshot. Newly published work is not incorporated during an experiment. In future work, we plan to equip the agents with live access to pre-print servers and citations graphs so that beliefs can adapt to emerging evidence in real time.
	\item \textbf{Limited evidence modalities}: We only retrieve textual passages. Important evidence such as figures, tables, or released code is ignored. We intend to extend the retrieval module to multi-modal documents and to incorporate program-of-thought execution for verifying quantitative claims.
	\item \textbf{Learned Refinement Policies}: The current refinement strategies (Deepening, Counterfactual, Hybridization) are based on pre-defined heuristics. We plan to learn a policy that selects refinement actions with reinforcement learning, using the same Bayesian utility as the reward signal.
\end{enumerate}

\section{Conclusion}

This paper presented HypoAgents, a Bayesian-entropy collaborative framework that enables a group of LLM-powered agents to generate, evaluate and refine research hypotheses in a closed loop. Experiments on 100 open-ended research questions from ICLR 2025 show that the framework boosts the average ELO score of hypotheses by 116.3 points after 12 iterations while simultaneously reducing uncertainty by 0.92. These results suggest that principled probabilistic reasoning, combined with information-theoretic exploration, offers a viable path toward reliable automated scientific discovery. Future work will address the outlined limitations.

\section{Acknowledgments}
This research was supported by the computational resources provided by our institution, which were essential for conducting the large-scale experiments presented in this work.

\bibliography{aaai2026}

\appendix

\section{Framework Architecture Diagram}

\begin{figure*}[htbp]
	\centering
	\includegraphics[width=0.9\textwidth]{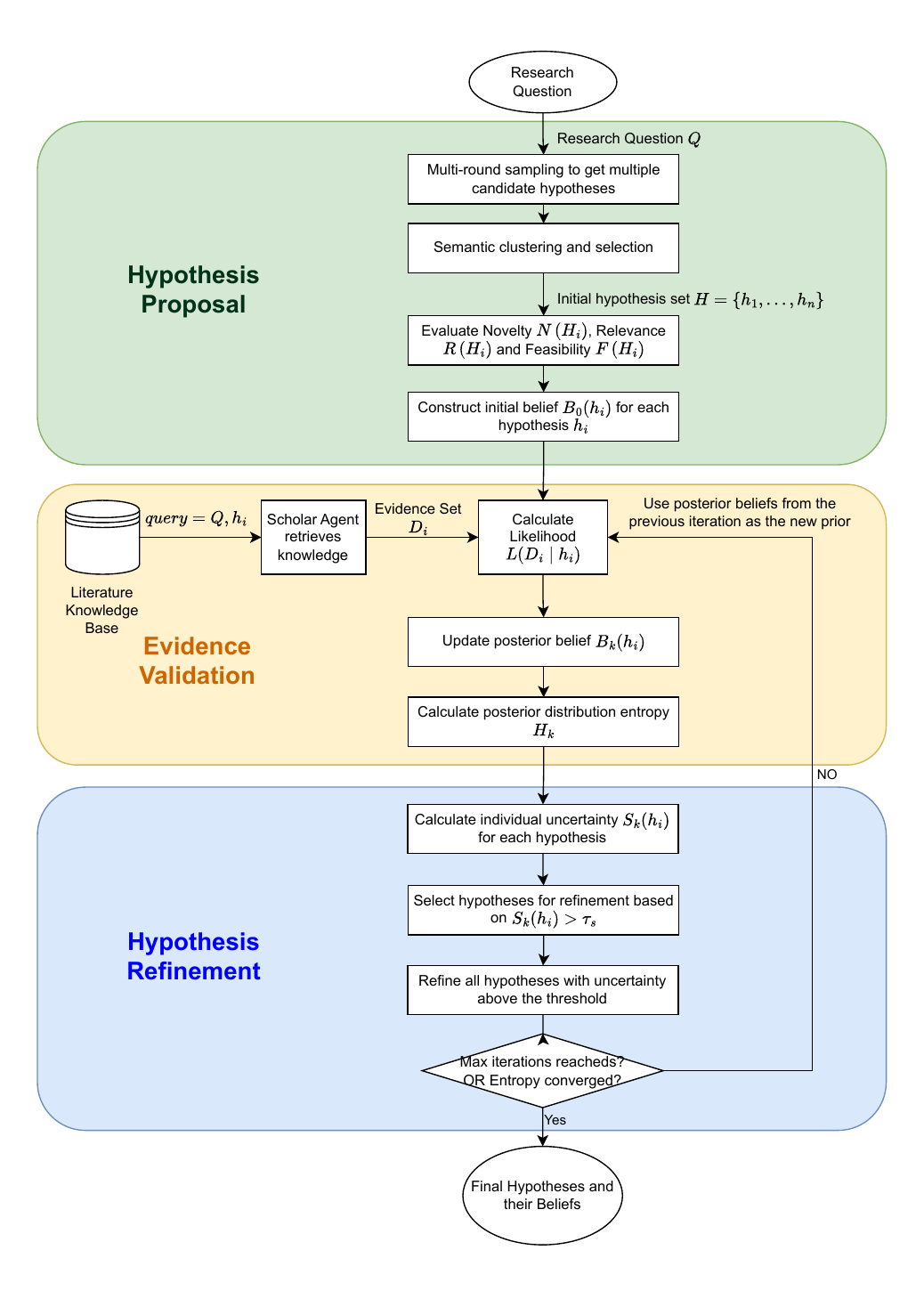}
	\caption{Flowchart of the \textbf{HypoAgents} framework}
	\label{fig:framework}
\end{figure*}

Figure~\ref{fig:framework} presents the flowchart of the \textbf{HypoAgents} framework, clearly depicting the interactions among its three core modules—hypothesis proposal, evidence validation, and hypothesis refinement—as well as the overall data flow within the system.

\section{Prompt Details}

\subsection{Prompt for Initial Hypothesis Generation}

\noindent\textbf{System prompt:}
\begin{quotation}
	You are an AI assistant specializing in academic research, particularly in artificial intelligence and machine learning. Your primary role is to assist researchers in formulating well-structured and theoretically grounded hypotheses, reviewing literature, and designing experimental methodologies.
	
	When helping with research hypotheses, ensure they are:
	
	1. Clearly framed within the current research landscape, identifying existing gaps.
	
	2. Grounded in strong theoretical foundations with relevant prior work.
	
	3. Precise and testable, specifying independent and dependent variables.
	
	4. Innovative and methodologically rigorous, distinguishing from existing approaches.
	
	5. Expected to contribute meaningfully to the research community.
	
	Always provide well-structured, concise, and publication-worthy responses. If clarification is needed, ask follow-up questions.
\end{quotation}

\noindent\textbf{User prompt:}
\begin{quotation}
	You are a senior research expert specializing in artificial intelligence. Your task is to propose a well-structured and theoretically grounded research hypothesis for a novel research problem that is suitable for publication in top-tier conferences and journals.
	
	Instructions for Generating the Research Hypothesis:
	
	Carefully analyze the given research question and develop a comprehensive, testable, and impactful hypothesis by incorporating the following key elements:
	
	1. Research Background \& Problem Statement:
	
	- Clearly describe the current state of research, existing challenges, and the core problem your hypothesis addresses.
	
	2. Theoretical Foundations:
	
	- Reference relevant prior work to justify the hypothesis, highlight unresolved gaps, and establish the rationale for your approach.
	
	3. Hypothesis Statement:
	
	- Formulate a precise and testable hypothesis, specifying the independent and dependent variables and their expected relationship.
	
	4. Methodology \& Innovation:
	
	- Outline the proposed research methodology, emphasizing the key ideas, novel contributions, and how it differs from existing approaches.
	
	5. Expected Contributions \& Impact:
	
	- Discuss the anticipated theoretical and practical contributions of the research, as well as its potential significance for the AI community.
	
	The research question you need to solve is: \texttt{research question}
	
	Carefully analyze the provided research question and construct a complete and coherent research hypothesis that meets high academic standards. The hypothesis should be written as a single, well-organized paragraph, ensuring logical flow and clarity. Avoid using bullet points, section headings, or Markdown formatting. Instead, provide a fluent and natural explanation as seen in top-tier research papers.
\end{quotation}

\subsection{Prompt for Novelty Evaluation}

\noindent\textbf{System prompt:}
\begin{quotation}
	You are a professor in the \texttt{keyword content} field, and you make a judgment about the novelty of a research hypothesis.
\end{quotation}

\noindent\textbf{User prompt}
\begin{quotation}
	You are an expert in an academic research field tasked with evaluating the novelty of a Hypothesis in the context of a given research Question. Novelty is defined as the degree to which a hypothesis is unique or innovative relative to existing knowledge or common methods. The background of existing knowledge can be inferred from the field of study and does not rely on specific documentary evidence.
	
	Follow these steps to evaluate:
	
	1. Understand the field and context of the research Question.
	
	2. Analyze whether hypotheses suggest new ideas, approaches, or mechanisms that have not been fully explored.
	
	3. Give a score (0–1) based on the degree of novelty, following the criteria:
	
	- \textbf{0}: not new at all, and highly coincident with common knowledge.
	
	- \textbf{0.5}: Medium novelty, partly based on existing knowledge but somewhat extended.
	
	- \textbf{1}: Highly novel, proposing a new perspective or approach.
	
	4. Briefly explain the reason for your rating (optional, but helpful).
	
	\medskip
	
	\textbf{Input:}
	
	Hypothesis: \texttt{\{hypothesis\}}
	
	Question: \texttt{\{question\}}
	
	\medskip
	
	\textbf{Output format (strictly adhered to):}
	
	\texttt{<novelty>\{your novelty rating\}</novelty>}
\end{quotation}

\subsection{Prompt for Likelihood Estimation}

\noindent\textbf{System prompt:}
\begin{quotation}
	You are a professor in the \texttt{\{keyword\_content\}} field, and you make a judgment about the likelihood of a research hypothesis.
\end{quotation}

\noindent\textbf{User prompt used to base likelihood estimation:}
\begin{quotation}
	Estimate the probability (0--1) that this evidence would be observed if the hypothesis is true.
	
	\medskip
	
	\textbf{Input:}
	
	Evidence: \texttt{\{knowledge\_content\}}
	
	Hypothesis: \texttt{\{hypothesis\}}
	
	\medskip
	
	Output format strictly adhered to: \texttt{<base\_LH>\{your match score\}</base\_LH>}
\end{quotation}

\noindent\textbf{User prompt used to methodology match:}
\begin{quotation}
	\textbf{Input:}
	
	Research question: \texttt{\{question\}}
	
	Evidence: \texttt{\{knowledge\_content\}}
	
	Hypothesis: \texttt{\{hypothesis\}}
	
	Check if the evidence contains methodologies supporting the hypothesis.
	
	Return 1 if matched, 0 if not, with a brief explanation.
	
	\medskip
	
	Output format strictly adhered to: \texttt{<match>\{your match score\}</match>}
\end{quotation}

\subsection{Prompt for Hypothesis Refinement}

\noindent\textbf{User prompt:}
\begin{quotation}
	You are an expert academic researcher specializing in hypothesis optimization within artificial intelligence research. Select ONE strategy below and state it exactly as shown:
	
	\textbf{A. Deepening}
	
	- Drill into causal mechanisms, define measurable variables, and tighten logical flow without adding new constructs.
	
	\textbf{B. Counterfactual}
	
	- Formulate the strongest plausible counter-hypothesis, rebut it with evidence, then revise the original hypothesis to survive the challenge.
	
	\textbf{C. Hybridization}
	
	- Import a concept or method from another discipline, integrate it with current evidence, and craft a hybrid hypothesis leveraging both domains.
	
	\medskip
	
	\textbf{Context}
	
	Research Question: \texttt{\{research\_question\}}
	
	Current Hypothesis: \texttt{\{hypothesis\}}
	
	Evidence Snippets (top 5): \texttt{\{evidence\_snippets\}}
	
	\medskip
	
	\textbf{Task}
	
	Using ONLY the chosen strategy, optimize the hypothesis to better address the research question while keeping it concise, precise, and empirically testable.
	
	\medskip
	
	\textbf{Output Format}
	
	Respond with a JSON object containing your optimized hypothesis and the optimization strategy:
	
	\texttt{\{}
	\texttt{"optimized hypothesis": "<your optimized hypothesis>",}
	\texttt{"strategy": "<A | B | C>"}
	\texttt{\}}
\end{quotation}

\section{Case Study}

To provide a concrete illustration of the HypoAgents framework's iterative optimization process, this section presents a case study. We trace the evolution of a single research hypothesis through several rounds of validation and refinement. The reader is encouraged to observe not only how the hypothesis text evolves from an initially broad formulation to a more precise and technically-grounded statement, but also the corresponding dynamics of two key metrics: the Belief Value (\(B_k\)) and the ELO Score. The Belief Value represents the framework's confidence in the hypothesis's validity, ranging from 0 (certainly false) to 1 (certainly true). Consequently, a belief value approaching 0.5 indicates maximum uncertainty, which marks the hypothesis as a prime candidate for refinement. This case study demonstrates how the framework, starting from a state of high uncertainty, progressively increases its belief as the hypothesis improves.

\paragraph{Research question}

\begin{quote}\itshape
	How can we efficiently adapt pretrained language models to domain-specific knowledge when facts appear rarely or only once in small corpora, overcoming the data inefficiency of traditional pretraining approaches?
\end{quote}

\paragraph{Round 1 Analysis}

\begin{itemize}
	\item Hyp ID: 1
	\item Belief Value: 0.437
	\item ELO Scores: 1339.86
\end{itemize}

\begin{quote}\itshape
	Pretrained Language Models (PLMs) have revolutionized natural language processing, exhibiting remarkable capabilities across diverse tasks; however, their effective adaptation to specialized domains typically necessitates extensive domain-specific corpora for continued pretraining or fine-tuning. This reliance on large datasets presents a significant challenge when domain-specific facts are rare or appear only once within small, proprietary corpora, leading to data inefficiency and sub-optimal knowledge acquisition. While existing methods like knowledge graphs and retrieval-augmented generation offer external knowledge access, the efficient *internalization* of novel, sparse facts directly into the model's parameters remains an unresolved problem, particularly for instances with minimal or single-occurrence evidence. Drawing inspiration from meta-learning principles that enable rapid adaptation, we hypothesize that a novel, gradient-based meta-learning framework, specifically designed to optimize for rapid knowledge acquisition and consolidation from extremely sparse, single-instance factual examples, will enable pretrained language models to efficiently integrate domain-specific facts from small corpora, demonstrating superior and more robust performance on knowledge-intensive tasks compared to conventional fine-tuning or continued pretraining approaches. This innovative methodology will involve training the model to learn an optimal 'update rule' or 'knowledge consolidation mechanism' that selectively prioritizes and encodes rare factual information during a few gradient steps, thereby overcoming the data inefficiency of traditional approaches and allowing the model to genuinely 'one-shot' or 'few-shot' learn specific facts directly into its parameters. Such a breakthrough would not only advance our theoretical understanding of efficient knowledge acquisition in large neural networks but also offer profound practical implications, enabling the rapid and cost-effective deployment of powerful PLMs in highly specialized, data-scarce domains—from niche scientific fields to proprietary corporate datasets—where precise factual recall from limited evidence is paramount, ultimately democratizing access to advanced AI for low-resource applications.
\end{quote}

\textbf{Analysis and Evaluation:} In the initial, the hypothesis text accurately identified the core of the research problem-the adaptability challenges of pretrained language models when dealing with small samples and sparse facts-and proposed a reasonable solution based on meta-learning. However, the formulation of this version more closely resembled the introduction of a research proposal than a concise scientific hypothesis. Its primary limitation was a lack of specificity: although it mentioned a "gradient-based meta-learning framework", it failed to articulate the specific mechanisms for its implementation or to explore the deeper theoretical problems it could address beyond data efficiency.

\paragraph{Round 5 Analysis}

\begin{itemize}
	\item Hyp ID: 4
	\item Binary Entropy: 0.817
	\item ELO Scores: 1252.91
\end{itemize}

\begin{quote}\itshape
	Traditional PLM adaptation to domain-specific knowledge struggles with rare, single-occurrence facts in small corpora due to data inefficiency and susceptibility to catastrophic forgetting. We hypothesize that a **meta-learning-driven, parameter-efficient adaptation framework** will overcome these limitations by **leveraging meta-learning's ability to \"learn to learn\" for rapid, few-shot adaptation** and **utilizing targeted parameter updates (e.g., via prompt tuning or adapters, modifying less than 1\% of parameters)**. This parameter-efficient approach will specifically encode new facts while preserving core model knowledge by preventing modification of general language understanding and reducing overfitting to sparse data, thereby mitigating catastrophic forgetting. This framework is particularly effective in low-resource scenarios and will enable PLMs to achieve superior factual recall and reasoning for scarce facts, demonstrating higher data efficiency, improved stability, and competitive or superior performance compared to traditional fine-tuning and continued pretraining.
\end{quote}

\textbf{Analysis and Evaluation:} After several rounds of iteration, the quality of the hypothesis underwent a qualitative leap, evolving into a highly refined and robustly argued scientific proposition. The improvements in this version clearly demonstrate the framework's "Deepening" optimization capability. Its progress is evident in two key aspects: \textbf{1) Proposing a specific technical path:} The hypothesis explicitly introduces a "parameter-efficient adaptation framework" and specifies implementation methods such as "prompt tuning or adapter", which significantly enhances the hypothesis's feasibility and verifiability. \textbf{2) Providing deeper theoretical support:} It precisely links the advantages of the parameter-efficient approach to solving the critical challenge of "catastrophic forgetting", indicating a more profound understanding of the problem. The text itself also became more concise and professional.

\paragraph{Round 8 Analysis}

\begin{itemize}
	\item Hyp ID: 4
	\item Binary Entropy: 1
	\item ELO Scores: 1284.60
\end{itemize}

\begin{quote}\itshape
	We hypothesize that a **meta-learning-driven, parameter-efficient adaptation framework** will efficiently adapt pretrained language models to rare, single-occurrence facts in small corpora. This framework leverages **meta-learning to acquire an optimal strategy for rapid, few-shot integration of novel, sparse factual patterns**, implemented through **targeted parameter updates (e.g., prompt tuning or adapters, modifying <1\% of parameters)**. This approach will specifically encode new facts while preserving general language understanding and mitigating catastrophic forgetting, thereby achieving superior factual recall, reasoning, data efficiency, and stability compared to traditional fine-tuning and continued pretraining in low-resource scenarios.
\end{quote}

\textbf{Analysis and Evaluation:} The hypothesis in this round can be regarded as the final refinement of a mature proposal. Its core ideas and technical path are largely consistent with the output of Round 5, with minor modifications primarily focused on sentence structure to improve fluency.

\paragraph{Abstract of the original paper}

\begin{quote}\itshape
	Pretraining on large-scale, unstructured internet text enables language models to acquire a significant amount of world knowledge. However, this knowledge acquisition is data-inefficient-to learn a fact, models must be trained on hundreds to thousands of diverse representations of it. This poses a challenge when adapting a pretrained model to a small corpus of domain-specific documents, where each fact may appear rarely or only once. We propose to bridge this gap with synthetic continued pretraining: using the small domain-specific corpus to synthesize a large corpus more amenable to learning, and then performing continued pretraining on the synthesized corpus. We instantiate this proposal with EntiGraph, a synthetic data augmentation algorithm that extracts salient entities from the source corpus and then generates diverse text by drawing connections between those entities. Synthetic continued pretraining with EntiGraph enables a language model to answer questions and follow generic instructions related to the source documents without access to them. If the source documents are instead available at inference time, we show that the knowledge acquired through our approach compounds with retrieval-augmented generation. To better understand these results, we build a simple mathematical model of EntiGraph, and show how synthetic data augmentation can \"rearrange\" knowledge to enable more data-efficient learning.
\end{quote}

\textbf{Analysis and Evaluation:} Comparing the framework's final generated hypothesis with the abstract of the original published paper reveals the capabilities and boundaries of current automated scientific discovery. A notable finding is that the high-quality hypothesis converged upon by HypoAgents (meta-learning + parameter-efficient fine-tuning) is methodologically distinct from the solution ultimately adopted by the human researchers (synthetic continued pretraining and the EntiGraph algorithm). This comparison yields two critical insights: 1. Autonomous Exploratory Capability of HypoAgents: The framework demonstrated its ability to autonomously conduct logical reasoning and exploration within a complex solution space, ultimately converging on a logically sound and highly valuable research proposition. This indicates the method's effectiveness in automatically generating high-quality, verifiable research ideas. 2. Boundaries of Current AI Creativity: Despite its impressive performance, HypoAgents primarily operates by combining and optimizing within known, established knowledge paradigms. In contrast, the human researchers exhibited "out-of-the-box" creativity by proposing a methodologically disruptive new paradigm. This highlights that "zero-to-one" disruptive innovation remains an irreplaceable core value of human researchers.

\end{document}